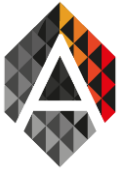

# Using multi-agent architecture to mitigate the risk of LLM hallucinations

Abd Elrahman Amer[a] and Magdi Amer[b]

[a] R&D, Java Innovations, 49 Burns Ave, Charlottetown, PEI, Canada, C1E 0B9

[b] ISG CMT Comms Practice, Cognizant, 300 Frank W Burr Blvd, Teaneck, NJ, United States, 07666

abdelrahman.amer@gmail.com, magdi.amer@cognizant.com

| KEYWORD | ABSTRACT |
|---|---|
| LLM; Gemini, Chat GPT, multi-agents, hallucination mitigation; Fuzzy Logic | Improving customer service quality and response time are critical factors for maintaining customer loyalty and increasing a company's market share. While adopting emerging technologies such as Large Language Models (LLMs) is becoming a necessity to achieve these goals, the risk of hallucination remains a major challenge. In this paper, we present a multi-agent system to handle customer requests sent via SMS. This system integrates LLM based agents with fuzzy logic to mitigate hallucination risks. |

## 1. Introduction

Recent advancements in Large Language Models (LLMs) have significantly enhanced the ability to develop systems that comprehend customer requests and determine the necessary actions to fulfill them. In today's competitive market, delivering superior customer service is crucial for attracting and retaining clients. Satisfied customers are more likely to become loyal, repeat buyers, and advocate for your brand, leading to increased revenue and market share (Strikingly, 2024).

In industries characterized by intense competition, implementing LLM-based services that effectively address customer needs and enhance satisfaction is becoming a key determinant of a company's growth and success. By leveraging LLMs, businesses can deliver more personalized, efficient, and scalable support, and thereby improve customer experience and foster loyalty (Iopex, 2024).

Furthermore, the integration of LLMs into customer service operations has been shown to reduce response times and increase efficiency. For example, Comcast's implementation of an LLM-based assistant for their agents resulted in a 10% reduction in time spent per customer interaction, translating to significant annual savings and improved customer satisfaction (Rome, Chen, Tang, Zhou, & Ture, 2024).

As the adoption of LLMs in customer service continues to grow, businesses that effectively utilize these technologies are likely to gain a competitive edge, leading to increased market share and long-term success.

One of the main drawbacks of LLMs that make industries hesitate before fully adapting LLM is the risk of hallucination. Hallucination occurs when a LLM presents information as facts when they are purely fictitious (Sun, Sheng, Zhou, & Wu, 2024). Recently, there was a legal case against a Canadian airline where the court held the airline reliable for the false advice that was given to the customer by the airline's chatbot (Garcia, 2024).

A similar case also occurred in the US, where a citizen received false advice from the chatbot of an insurance company. The customer acted based on this advice, but his claim was denied later. After the customer reached out to KSL News, which contacted the insurance company on the customer's behalf, the company decided to honor the advice of the chatbot and pay the customer the promised sum (Gephardt, 2024) . Moreover, there were several cases in the U.S. and Canada where lawyers presented to the court legal documents that were generated using LLM. Upon examination, it was discovered that the documents referred to fabricated case laws, resulting in legal responses ranging from warnings to sanctions (Ogunde, 2024).

This motivated the research presented in this paper. We are presenting a multi-agent architect that aims to reduce the cost of operation in addition to mitigating the risk of hallucination.

In the following section, an overview of LLM usage as an intelligent agent and using multi-agents will be presented. Next, the proposed architecture will be explained followed by the discussion of the implementation details and evaluation. Finally, the conclusion and future work will be presented.

# 2. Background

In (Franklin & Graesser, 1996), the authors identified the set of features that differentiate an agent from other programs. These features include pro-activeness, being autonomous, the ability to perceive its environment, being goal oriented and the ability to communicate with other agents.

The real inspiration of intelligent agents came from the observation of the behavior of ants and bees. In fact, the analogy between a swarm of ants and bees has long been the inspiration behind multi-agent architecture (Lemmens, de Jong, Tuyls, & Nowe, 2008). Let's take a beehive for example. Each individual bee has an extremely specific and fixed role, such as the queen bee or the scout bee. Each role requires limited training and limited intelligence. The collective intelligence of the beehive that comes from their collaboration exceeds the sum of the intelligence of individual bees. In fact, the bee, despite its small brain that consists of around 1 million neurons, can exert collective intelligence that scientists thought was exclusively unique to humans (Menzel, 2012) and (Bridges, et al., 2024). Interestingly, studies conducted on bee hives (Seeley, Passino, & Visscher, 2006) have proved that the hive does not take decisions based on the opinion of an individual bee, but instead multiple scout bees are used to confirm the assessment of a scout bee.

In (Xi, et al., 2025) the authors confirmed that LLMs are exhibiting many capabilities that are expected from a software agent, making them suitable to construct agent-level intelligence. Furthermore, the use of tools allows the LLM-agents to overcome inherent limitations of LLMs. In fact, LLMs are not capable of responding to current events, such as today's weather or stock price. Using tools allows LLMs to delegate specific tasks to these tools, which may be used to retrieve current information or to fetch data related to a company's products (Li, et al., 2023).

In (Guo, et al., 2024), the authors confirmed the suitability of using LLM to build multi-agent applications, but they affirmed that such applications have an increased risk of hallucinations, as one agent's hallucination may have a cascading effect.

The authors also identified 4 communication models that are used for agent communication: Layered, centralized, decentralized and shared messages pool. In layered communication, agents are organized into layers, where each agent can only interact with the agents at the same layer or adjacent layers. In centralized communication, agents communicate with a centralized agent that usually arbitrates between the agents and that makes decisions based on their findings. Decentralized agents allow agents to communicate directly with each other, but such architecture requires each agent to be aware of other agents in the ecosystem, and to implement specific protocols to communicate with each agent. Conflict detection and resolution are also more challenging in decentralized architecture. In a Shared Message Pool (Hong , et al., 2024), agents publish messages in a message pool and subscribe to messages they are interested in. This communication model provides complete decoupling between agents, System resilience is achieved through agent availability and the decoupling of message throughput, allowing agents to produce and consume messages independently of the performance of other agents in the system. In our research, we adapted the Shared Message Pool using the dynamic factory model, as will be explained in the following section.

There are multiple research papers that are focused on mitigating hallucinations in LLM. Sentence-BERT (Reimers & Gurevych, 2019) improved Cosine similarity calculation between two sentences by using a fixed-size vector representations of sentences. This is achieved using a Siamese or triplet network structure with shared BERT weights and a pooling layer. This provides accuracy similar to that of BERT, while achieving performance improvement by several orders of magnitude.

In (Fu, Ng, Jiang, & Liu, 2023), the authors introduced **GPTScore**, a framework that evaluates LLM-generated text based on 22 qualitative aspects such as Fluency, Coherence, Relevance, and Specificity. This is achieved by prompting the language model with natural language questions for each aspect (e.g., "How well is the generated text relevant to its source text? Rate from 1 [not at all] to 5 [perfectly]."). The model's response probabilities are then used to compute scores, and the final evaluation score is derived using a weighted aggregation function. GPTScore enables flexible, instruction-based evaluation without the need for supervised training data. The proposed system's main drawback is the computational cost required to evaluate a single answer.

In (Ji, et al., 2023), they classified hallucination errors in 3 categories; Fact Inconsistency, Query Inconsistency, and Tangentiality. Fact Inconsistency involves answers that contradict known facts, typically due to failure in recalling relevant knowledge. Query Inconsistency describes outputs that are unrelated or nonsensical in relation to the input query, often lacking any meaningful connection to the question. Tangentiality, while not classified as hallucination, refers to answers that remain topically relevant but do not directly address the query due to insufficient reasoning or inference.

To address the issue of hallucination, the authors proposed to use a self-reflection process, consisting of 3 loops; Factual Knowledge Acquiring Loop, Knowledge-Consistent Answering Loop, and Question-Entailment Answering Loop. In the first loop, the system requires the LLM to provide background knowledge related to the question being asked. Using the scoring system developed by (Fu, Ng, Jiang, & Liu, 2023), the LLM's answer is evaluated. The system will require the LLM to refine the answer if the score is below a pre-defined threshold, hence providing the first loop. If it exceeds the threshold, it goes to the second part of the system, where the system will ask the LLM to answer the question based on the knowledge gathered in the first step. Like the first step, the answer is evaluated and the LLM is required to refine the answer if its evaluation is below the threshold. Otherwise, the answer is evaluated via Sentence-BERT (Reimers & Gurevych, 2019) to calculate the cosine similarity between the question and the response of the system. If the answer does not meet the satisfactory entailment level, the process is repeated from the beginning, otherwise the answer is presented to the user. The approach described in this research shows that the proposed architecture reduces hallucinations in the domain of Medical Generative Question Answering. It cannot be applied to the problem of interpreting the request of a client, which is the focus of this paper.

In (Darwish, Rashed, & Khoriba, 2025), the authors built a multi-agent architecture to reduce the probability of hallucination when summarizing the calls of a support center. This is achieved by using a Consulting agent that performs the summarization of the text and a Reviewer agent that evaluates the quality of the summarization using Gestalt similarity matching. This loop may be repeated up to 3 times if the pattern matching is below a certain threshold. Using a sample of 3680 chunks in 308 transcripts, the system was able to achieve a maximum similarity score of 89.56% after three iterations when using LLaMA3-3-8B. The advantage of this approach is that it does not require training and that it is not domain specific.

Both (Fu, Ng, Jiang, & Liu, 2023) and (Darwish, Rashed, & Khoriba, 2025) propose valuable architectures to mitigate hallucination risks in the domains of question answering and summarization, respectively. However, their approaches are not applicable to the challenge of interpreting instructions received via customer SMS messages, which is the focus of this research.

## 3. Problem Description and Solution Architecture

In this paper, we are proposing a multi-agent architecture that uses a combination of fuzzy logic and LLM to improve the ability to detect LLM hallucinations when interpreting customer requests received via SMS messages and mitigate such risks.

We have chosen a scenario from the industry to implement our proof of concept. In this scenario, a pharmacy sends automated messages to customers, prompting them to reply with a specific code to either renew an expiring prescription or request that it not be renewed. Different medications will be assigned to different codes. Customers may combine multiple codes in a single message to instruct the pharmacy which medications they want to renew and what others they want to stop. In addition to these codes, customers often include sentences to provide specific instructions or ask questions. Using traditional technologies would fail to process these messages, as the application would be incapable of interpreting the customer's requests. Using the LLM ability to understand natural language would be suitable to improve the success rate of message processing.

The first component of the architecture is the Incoming SMS service (1) as shown in **Figure 1**. This service is responsible for authenticating the user and for placing the SMS message of the customer in the incoming SMS event hub (2). Each message has a set of meta data, including the event id, which is a unique identifier for the message, and the type of message. In this paper, we are using "renewal" messages, which are the message that the customers send to the system to reply to renewal messages. These messages will be consumed by the Orchestration Agent (A), which contains several components; the Dynamic Event Processor (3), the Dynamic Agent Event Services (4), the Processing Service Dispatcher (5), and the Orchestration Worker Agents (6).

The Dynamic Event Processor (DEP) (3) reads the configuration file and dynamically creates Dynamic Agent Event Services (DAESs) (4) for each set of conditions. The role of the DAE is to inspect the attributes of the message to determine if they match those expressed in the conditions. If there is a match, they will request the Processing Service Dispatcher (PSD) (5) to provide them with the corresponding Orchestration Worker Agent (OWA) (6) that is responsible for handling messages of this type. Multiple DAES may match the same message; in which case the message processing will be done in parallel using Spring Boot and WebFlux. **Figure 2** shows the configuration used to create the DAEs.

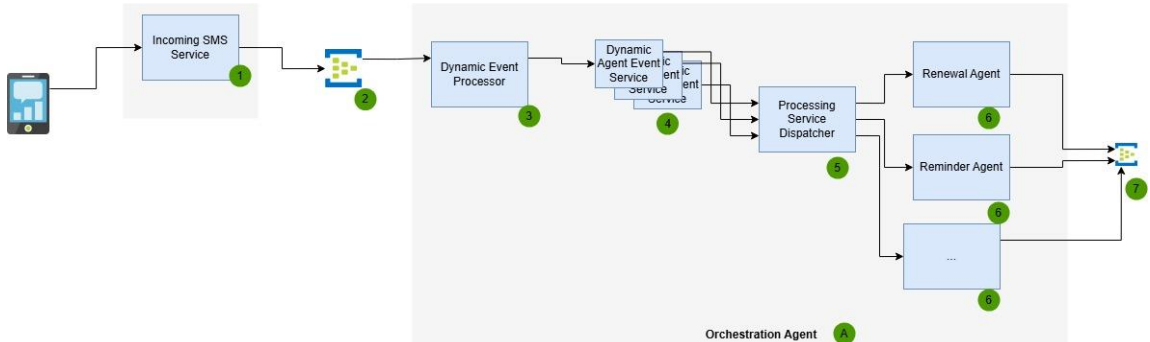

**Figure 1:** *Architecture overview of the SMS events flow from the customer's mobile to the corresponding Agent.*

```
services:
  rules:
    - name: DynamicServiceA
      qualifier: RenewalAgent
      conditions:
        - key: "metadata.type"
          value: "renewal"
    - name: DynamicServiceB
      qualifier: ReminderAgent
      conditions:
        - key: "metadata.type"
          value: "reminder"
```

**Figure 2:** *Configuration for creating event Service Agents Dynamically. Here, two Event Service Agents will be created. The first will subscribe to events whose types are renewal, while the second will subscribe to reminder events. The first and second agents will forward the matched events to the Renewal-Agent and Reminder-Agent, respectively.*

One of the main advantages of using multi-agent architecture is that the technology used to implement the intelligence of each agent may differ, hence allowing the designer to choose the AI implementation that is more suitable to solve the problem for which the agent is responsible (Goonatilleke & Hettige, 2022).

In this paper, we will focus on the one type of Orchestration Worker Agent, which is the Renewal Agent (6). The Renewal Agent is responsible for handling Renewal Messages. Agents responsible for other messages may follow the same approach that is implemented by the Renewal Agent.

The Renewal Agent (RA) implementation uses regular expression and Fuzzy Logic. The RA connects to the database and caches the list of keyword pairs that may be used by the customer to express whether they want to renew or stop a medication. It also caches the regular expression used to identify these keywords and the variations allowed in these keywords. It then removes the known politeness-expressions, such as "please" and "thank you." The RA splits the message into individual words. Next, the RA tries to match each word in the message with the regular expressions of the keywords. If all the words of the message were identified as keywords, this means that the RA was able to fully process the message. A Fuzzy variable, called degree of confidence, will be created to reflect how confident the agent is that it was able to fully understand the message. The degree of confidence has 3 labels: high, intermediate, and low. They are calculated based on the percentage of words that were identified vs the total number of words in the message. The fuzzy logic implementation is done using JFuzzyLogic (Cingolani & Alcalá-Fdez, 2012). If the RA was able to identify 100% of the keywords, then the high-degree-of-confidence will be 1, with the other two labels zero. If some of the keywords were not identified, then the high, intermediate, and low confidence will be calculated accordingly.

```json
{
  "metadata":{
    "type":"renewal",
    "eventId":"A1010",
    "customerId":"C1001",
    "stepId":"S001",
    "customerEventTime":"2025-01-15T10:48:46Z",
    "lastUpdateTime":"2025-01-15T10:49:08Z"
  },
  "renew":[
    "keyword1",
    "keyword2"
  ],
  "stop":[
    "keyword3"
  ],
  "degreeOfConfidence":{
    "high":0.85,
    "intermediate":0.6,
    "low":0
  }
}
```

**Figure 3:** *The Json Message that the Renewal Agent will post in the Agent Communication Topic. This message contains the meta data section about the creation time of the SMS, the customer id that has sent this message, the step the processing of the message is in, and the list of 'renew' keywords and 'stop renew' keywords that were detected by the Renewal Agent. It also contains the degree of confidence, which is a fuzzy variable that reflects how confident the agent is in its ability to fully understand this message.*

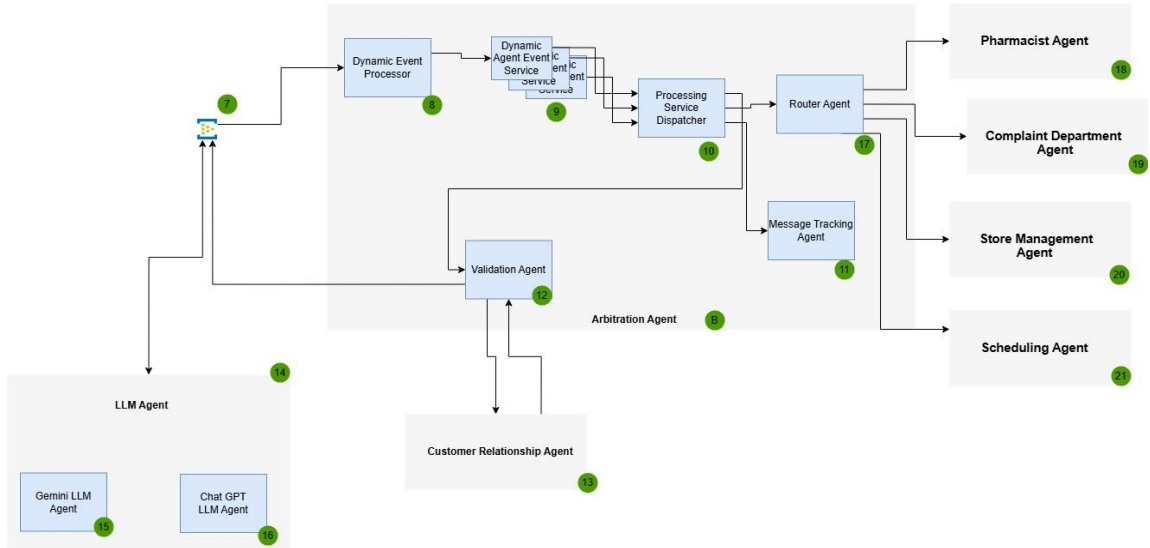

**Figure 4:** *Overview of the Message Flow within the Arbitration Agent and the Relationship with the Downstream Agents for Processing the Patient SMS.*

The RA (Reasoning Agent) will publish the processed result of the event to the event hub's topic designated for Inter-Agent Communication (7), which will hereafter be referred to as the agents' topic. Figure 3 illustrates the JSON structure of the message after RA processing. Additionally, the RA stores the original message in the database, using the event ID as a unique identifier. The event hub facilitates a shared pool communication model, as described by (Guo, et al., 2024). Upon completing its processing, the RA assigns a step ID value of '1' to indicate that the initial handling is complete and that a subsequent decision regarding the message is required.

The Arbitrator Agent (B), shown in **Figure 4**, contains several components. The DEP (8), DAESs (9) and PSD (10) play a similar role as the counterparts in the Orchestration Agent. They allow us to select which agent is suited to handle the event in the agents' topic. More than one agent may be selected to handle the same event in parallel. In fact, the Message Tracking Agent (MTA) (11) will always be selected in addition to other agents. MTA's role is to store in the database the details of each step a message went through, hence providing the ability to track the status of each message.

When the RA writes a message to the agents' topic, it will set the state-id to 1. The PSD will forward the message to the Evaluator Agent (EA) (12). The EA will base its decision on how to handle the message based on the fuzzy variable degree-of-confidence that was set by the RA, in addition to the fuzzy variable customer-importance that is set by the Customer Relationship Agent (CRA) (13). The CRA converts the years since the customer joined and the total amount of purchases that the customer made in the last 12 months into fuzzy variables and uses fuzzy rules to compute the fuzzy variable reflecting the customer-importance.

The EA begins by examining the value of the high-degree-of-confidence of the message that was set by the RA. If this value is 100%, this means that the RA was able to fully understand the message. This will occur when the customer's SMS contains only keywords and simple politeness-expressions. In this case, each keyword will be sent via rest-API to the Pharmacy Service Endpoint to cancel or renew the medications, as instructed by the customer. No further processing is needed, and the processing of the message is completed.

```
RULEBLOCK No2
    AND : MIN;      // Use 'min' for 'and' 'max' for 'or'
    ACT : MIN;      // Use 'min' activation method
    ACCU : MAX;     // Use 'max' accumulation method
RULE 7 : IF customerImportance IS high  THEN action IS forwardToLLM;
RULE 8 : IF degreeOfConfidence IS high THEN action IS forwardToLLM;
RULE 9 : IF customerImportance IS medium AND degreeOfConfidence is medium
THEN action IS forwardToLLM;
RULE 10 : IF customerImportance IS medium AND degreeOfConfidence is low
THEN action IS fail;
RULE 11 : IF customerImportance IS low AND degreeOfConfidence is medium
THEN action IS fail;
RULE 12 : IF customerImportance IS low AND degreeOfConfidence is low THEN
action IS fail;
END_RULEBLOCK
```

**Figure 5:** *The fuzzy rules governing the choice of the Evaluator Agent. The Rule-Block determines the action that will be chosen by The Evaluator Agent.*

```
{
    "renew": [
        "keyword1",
        "keyword2"
    ],
    "stop": [
        "keyword3",
        "keyword4",
        "keyword5"
    ],
    "complaint": [
        "first complaint issue",
        "second complaint issue"
    ],
    "request": [
        "first request",
        "second request",
        "third request"
    ],
    "mood": "positive"
}
```

**Figure 6:** *Defining the output format for the LLM Agent. The response will contain an array of keywords that are used to renew a medication and another array for the keywords to stop medication renewal. It also contains a list of complaints and requests that the customer may have included in the message with the evaluation of the mood of the message.*

If the high-degree-of-confidence of the message is less than 100%, this means that the SMS contained a few words and sentences that the RA was not able to match with keywords. Depending on the fuzzy rules shown in **Figure 5**, the EA will decide to either forward the message to the LLM agent or fail the message. Message failure will result in an SMS being sent to the customer, asking them to call customer support.

Depending on the output of the fuzzy rule system, the EA may decide that the message should be processed by the LLM Agent (LA) (14). In this case, the EA will write to the agents' topic a new message with step-Id 2. The Kafka client of the LA will consume all messages whose step-id is equal to two. It will use LangChain4J (Goncalves, 2024) to communicate with the LLM. LangChain4J provides a layer of abstraction hiding the details of the LLM being called, allowing us to call both Gemini (Google, 2025) and ChatGPT (OpenAI, 2025).

Using prompt engineering and a few shots approach, the LLM reply classifies the list of keywords. It also extracts from the SMS message the list of complaints and the list of requests, as shown in **Figure 6**.

The Validator Agent (VA) will subscribe to the event hub, consuming messages from Chat GPT and Gemini. The key role of the VA is to evaluate the probability that the response of the LLM Agents was the result of hallucination. This is done in two stages. In the first stage, the VA inspects the keywords extracted. In the second stage, it inspects the list of complaints and requests extracted.

To detect hallucinations in keyword extraction, the VA will compare, for each response, the keywords extracted by the RA and those identified in the LLM response for the response obtained from Gemini Agent and Chat GPT Agent. If the RA was able to extract keywords that are not present in the response of the LLM Agent, then this means that the LLM response was subject to hallucination. This is due to the fact that RA uses parsing techniques guaranteeing any extracted keywords to be 100% correct. RA weakness is that it may not be able to extract all the keywords in the SMS message. If the LLM agent was able to detect a keyword that is not present in the original SMS text, this also means that the response is the result of a hallucination. If the responses obtained from Gemini and Chat GPT were not discarded and the keywords extracted by both of them don't match, the request will be forwarded to both of them one more time. If the response still doesn't match, an SMS will be sent to the patient requesting them to contact customer support.

If the LLM Agent was able to identify keywords more than that identified by the RA agent, we will inspect to see if that keyword is related to a stop request, in which case, the VA agent will apply the set of fuzzy rules to determine the value of high-risk depending on the type of medication, how long the patient has been taking this medication, whether it is associated with a chronic or long-term disease. Using the center-of-gravity defuzzification technique, the VA agent will calculate the crisp value of the high-risk. If it exceeds the threshold parameter as defined in the Spring Boot config-server, the VA will send an SMS to the patient requesting confirmation. Otherwise, the list of keywords will be processed by sending each keyword to the rest-API to the Pharmacy Service Endpoint, as explained previously. The activity diagram describing this stage is shown in **Figure 7**.

The second stage is to confirm that the list of complaints and requests is not the result of hallucinations. This is achieved by requiring the Gemini LLM to evaluate the response of the ChatGPT LLM and vice versa. To do that, the original SMS text and the section related to the complaints and requests are sent to the LLM, requesting it to provide a score from 1 to 10. A score of 10 indicates a high accuracy, while a score of 1 indicates a high level of inaccuracy. An evaluation of less than 5 would be interpreted as a bad interpretation and will be discarded. If both responses are discarded, the client will be required to contact Customer Support, otherwise the response with the highest score will be chosen. This technique uses the same concepts explained in (Fu, Ng, Jiang, & Liu, 2023) and (Darwish, Rashed, & Khoriba, 2025). In a future extension of this research, more advanced techniques for improving the quality of the response will be implemented.

Each request and complaint will be sent to the Router Agent. Several "Expert Agents" register with the Router Agent, each indicating their area of expertise and the problems they are best suited to handle. The Router Agent will choose the Expert Agent to handle a given request or complaint using prompt engineering and will forward the request or complaint to the corresponding agent. In our research, we defined four Expert Agents: the pharmacist agent, the store management agent, the scheduling agent, and the complaint department agent. **Figure 8** shows the response of the router agent to the complaint "bad taste of medicine". The response has chosen to send the complaint to the pharmacist agent. It also rephrased the complaint to better express the concept the client wanted to convey.

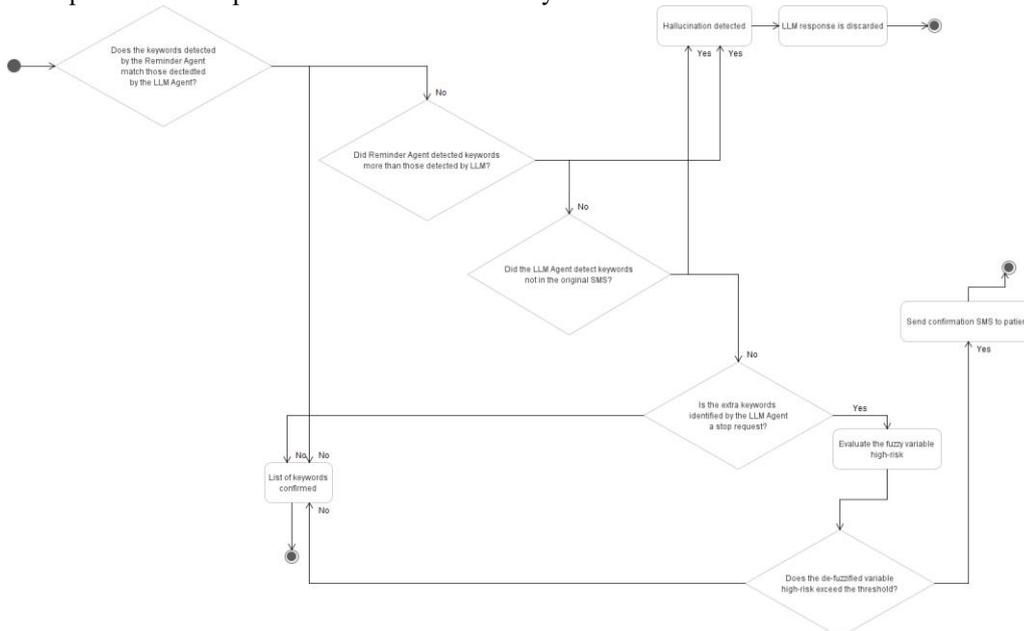

**Figure 7:** *Activity Diagram Explaining the Process of Mitigating the Risk of Hallucination in the List of Keywords by the Validation Agent.*

```
{
    "destination": "Pharmacy",
    "next_inputs": "I have a complaint about the bad taste of a medication."
}
```

**Figure 8:** *Response of the Router Agent to the Complaint "Bad Taste of Medicine". The agent has chosen the expert agent that is best suited to handle the complaint. It also rephrased the complaint to improve the sentence quality.*

The expert agents may be implemented using different technology. The Pharmacist Agent will forward the message to the Pharmacist to handle questions and requests from patients. Similarly, the complaint department agent will forward the message to customer-support to handle the complaint.

The store management agent is a RAG agent, which is suitable to answer questions related to store locations, normal operation hours, and holiday-hours of operation.

The scheduling agent is used to book appointments for vaccinations. It uses LLM model augmented with tools (Goncalves, 2024) to link the LLM model to the Appointment Tool. The request to book an appointment will be sent to the LLM, which will interpret the patient's constraints and preferences, and generate a list of suitable time slots. This list is then sent to the Appointment Tool. The Appointment Tool will query the database and attempt to select an available appointment that meets the patient's request. The LLM will use the tool's response to formulate a reply to the original query.

For example, if the request "I want an appointment on Saturday 03/22/2025 afternoon or Sunday 03/23/2025 morning" is sent to the LLM, it will forward to the Appointment Tool a list of suggested slots in the format: "year: 2025, month: 6, day: 29, hours: 10." The tool will select a time and return it to the LLM, which will respond with a message such as: "Answer: You can schedule your appointment for Saturday, June 28, 2025, at 5 PM." This suggested time will then be processed, and the patient will receive an SMS confirmation.

## 4. Result

For our proof of concept, we tested the system using ten sample SMS messages in the event hub. These sentences varied in structure and tone in order to evaluate keyword extraction. Below are the ten set of messages used for the test:

1. "1, unenroll. Thank you for your great service. I just want to know if it is normal for the medication to taste so bad? I also noticed that my blood pressure medication has no renewal left. Do I need to call my doctor or can you renew it for me?"

2. "1, unenroll. Thank you"

3. "Could you please execute the following. 1"

4. "Thank you for your great service. 1, renew."

5. "no, 2"

6. "Could you please do 1"

7. "Stop. I want to reserve a reservation for the flu vaccine"

8. "1 But this time I want to get the medication by next week"

9. "Enroll. I want also to renew my blood pressure medication.'

10. "Please stop sending me text messages"

Both the RA and the LLM agents were able to successfully extract all relevant keywords from the messages, with exceptions arising only when keywords appeared in the middle of complex or multi-intent sentences, such as in the last two examples. In example 9, the RA had extracted the keyword "enroll" and not the false hit "renew". As this word is not related to a stop keyword, it was considered low-risk and was sent to the pharmacy for processing. In the second case, the RA didn't extract any keywords, but the LLM agents extracted "stop". Depending on the medication associated with that keyword, the system will either process the request or decide to send a confirmation message to the patient.

Another test we conducted was to send the first question 50 times. The keywords extracted from both agents were correct in all cases except once. In this instance, one of the LLM agent's responses was {'renew': [ '1', 'renew' ], 'stop': [ 'unenroll' ]}. In this case, it wrongly added an extra keyword "renew" that was not part of the original text of the SMS message. Hence that response was discarded by the VA agent and the response by the other LLM agent was used to process the message.

Based on the set of tests conducted, our results indicate that the system performs as intended—successfully identifying certain hallucination instances, assessing hallucination risk, and applying appropriate mitigation strategies based on hallucination type. A comprehensive assessment of the system's success rate, however, will require deployment and evaluation in a production environment, which remains outside the current scope of this study.

## 5. Conclusion

In this paper, we presented a proof of concept for a multi-agent architecture designed to handle incoming SMS messages from patients. Message processing is decomposed into sub-problems, with each sub-problem handled by a specialized intelligent agent. The intelligence of each agent is implemented using either LLM, parsing techniques, or Fuzzy Logic, depending on which approach is most suitable for the task. Keyword hallucination is detected by comparing the outputs of the LLM Agent and the Parser Agent. Risks are evaluated, and decisions are made based on the fuzzy rules defined in the Validation Agent. The extraction of requests and complaints employs a feedback validation approach to minimize the risk of hallucination. In a future extension of this paper, we plan to explore more advanced techniques to further improve the quality of extracted complaints and requests.